\documentclass[10pt,twocolumn,letterpaper]{article}

\usepackage[pagenumbers]{cvpr} 

\usepackage{graphicx}
\usepackage{amsmath}
\usepackage{amssymb}
\usepackage{booktabs}
\usepackage{comment}
\usepackage[accsupp]{axessibility}

\usepackage[pagebackref,breaklinks,colorlinks]{hyperref}

\usepackage[capitalize]{cleveref}
\crefname{section}{Sec.}{Secs.}
\Crefname{section}{Section}{Sections}
\Crefname{table}{Table}{Tables}
\crefname{table}{Tab.}{Tabs.}

\def\cvprPaperID{8133} 
\def\confName{CVPR}
\def\confYear{2022}
\def\JW#1{{\color{blue}{\bf[JW:}{\it{#1}}{\bf ]}}}
\newcommand{\LT}[1]{\textcolor{red}{{\bf LT:} #1}}

\begin{document}

\title{Deformable Video Transformer}

\author{Jue Wang$^{1}$
\quad
Lorenzo Torresani$^{1,2}$\\
$^{1}$Facebook AI Research \quad $^{2}$Dartmouth \quad\\
}
\maketitle

\begin{abstract}
Video transformers have recently emerged as an effective alternative to convolutional networks for action classification. However, most prior video transformers adopt either global space-time attention or hand-defined strategies to compare patches within and across frames. These fixed attention schemes not only have high computational cost but, by comparing patches at predetermined locations, they neglect the motion dynamics in the video. In this paper, we introduce the Deformable Video Transformer (DVT), which dynamically predicts a small subset of video patches to attend for each query location based on motion information, thus allowing the model to decide where to look in the video based on correspondences across frames. Crucially, these motion-based correspondences are obtained at zero-cost from information stored in the compressed format of the video. Our deformable attention mechanism is optimized directly with respect to classification performance, thus eliminating the need for suboptimal hand-design of attention strategies. Experiments on four large-scale video benchmarks (Kinetics-400, Something-Something-V2, EPIC-KITCHENS and Diving-48) demonstrate that, compared to existing video transformers, our model achieves higher accuracy at the same or lower computational cost, and it attains state-of-the-art results on these four datasets.
\end{abstract}
\section{Introduction}
Although transformers~\cite{vaswani2017attention} were originally proposed to address NLP problems, they have quickly gained  popularity in computer vision after the introduction of the Vision Transformer (ViT)~\cite{dosovitskiy2020image}. Compared to CNN architectures, which can model pixel dependencies only within the small receptive fields of convolutional filters, ViTs offer the benefit of capturing longer-range dependencies. This is achieved by means of the self-attention operation,  which entails comparing features extracted from image patches at different locations. The downside of self-attention is that it causes a high computational cost if executed globally over all pairs of image patches. In fact, the complexity of global self-attention is $\mathcal{O}(S^2)$ where $S$ is the number of patches, which can be in the thousands. 

\begin{figure}[]
\centering
  \includegraphics[width=1.0\linewidth]{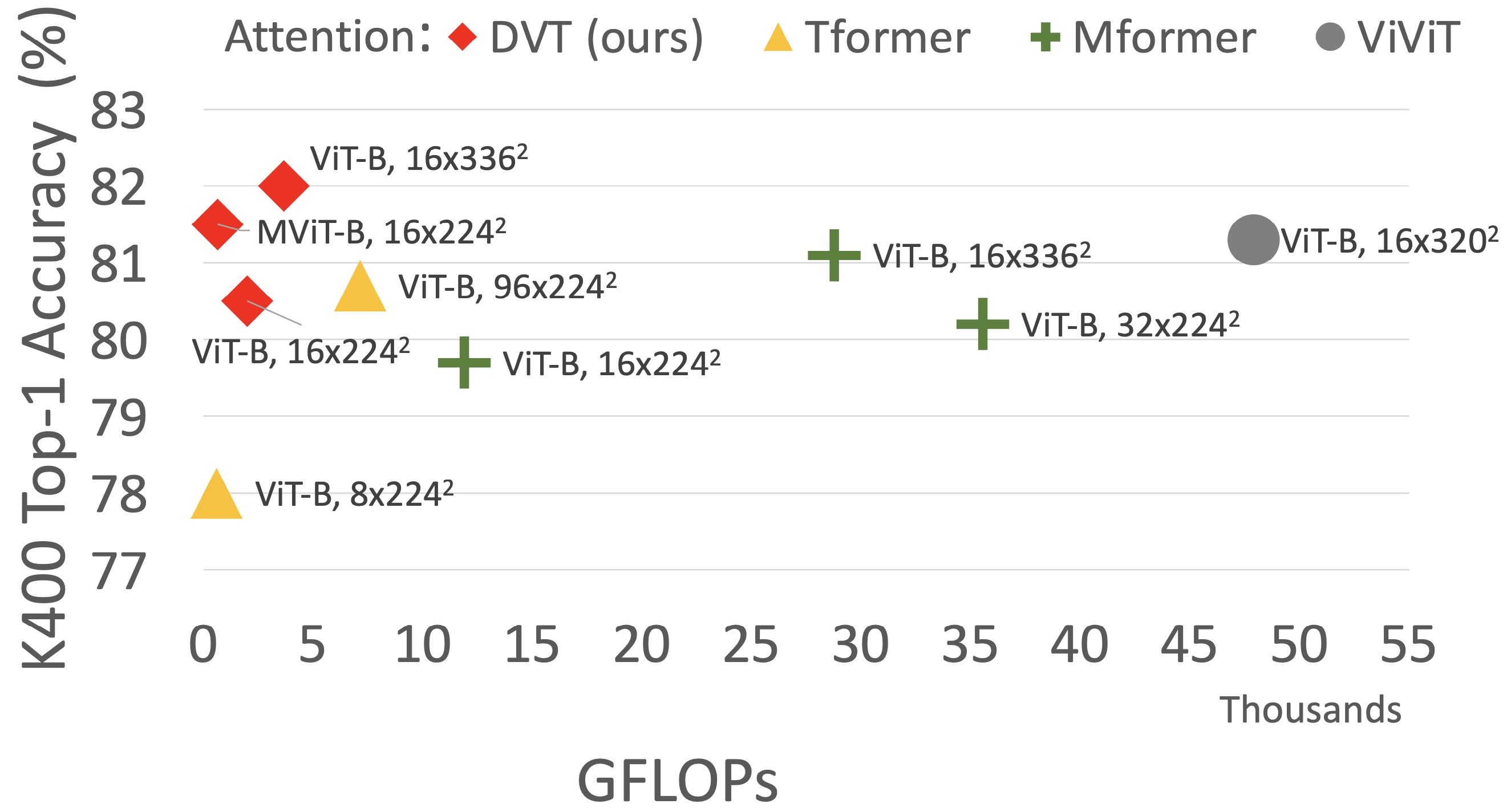}
\caption{Computational cost vs classification accuracy on K400 for video transformers pretrained on ImageNet-21K. Red diamonds denote models employing our proposed attention scheme (DVT). Our method achieves consistently better accuracy and lower computational cost compared to prior attention strategies (Tformer~\cite{bertasius2021space}, ViViT~\cite{arnab2021vivit}, Mformer~\cite{patrick2021keeping}). Text tags next to markers specify video transformer architectures and clip sizes.}
\label{fig:gflop_k400}
\end{figure}

In the case of videos, global self-attention becomes especially costly, since the number of patches to consider scales linearly with the number of frames $T$ in the clip. Thus, the cost becomes quadratic in the space-time resolution, i.e., $\mathcal{O}(S^2T^2)$. Exhaustive comparison of all space-time pairs of patches in a video is not only prohibitively costly but also highly redundant, since adjacent frames are likely to contain strongly correlated information. One way to reduce the computational cost is to limit the temporal attention to compare only pairs of patches in different frames but at the {\em same} spatial location~\cite{bertasius2021space, arnab2021vivit}. The complexity becomes $\mathcal{O}(S^2 T)$. However, this form of attention forces the model to perform temporal reasoning by looking through 1-dimensional ``time tubes'' of individual patches. Unfortunately, due to camera and object motion, the scene region projected onto a particular patch in frame $t$ may be completely unrelated to the scene information captured in the same patch at a different time $t'$. For example, the object appearing in the patch at time $t$ may have moved outside that patch in frame $t'$, thus rendering the recovery of motion information impossible and making the comparison between the two patches less relevant. Some works have proposed pooling tokens from local windows~\cite{fan2021multiscale}, or merging patches and limiting the self-attention operation to local space-time neighborhoods~\cite{liu2021video}. While effective in reducing the computational cost of video transformers, these methods employ fixed, hand-designed attention strategies that compare patches at predetermined locations, ignoring the motion in the video. 

In this work we propose a novel space-time attention mechanism for video transformers, which we name {\em deformable space-time attention}. While previous works rely on fixed schemes of attention which neglect the dynamic nature of the video, our method leverages motion cues to determine which patches to compare, thus implementing a form of input-conditioned attention. Given a query patch at a certain space-time location, our method samples $N$ patches to compare to that query patch in each of the other frames. The locations of these $N$ samples are predicted according to motion cues relating the query patch to the frame to attend. In other words, our strategy decides dynamically ``where to look'' in each frame on the basis of the appearance of the query patch and its estimated motion. Importantly, the motion cues are obtained at zero cost directly from the information stored in the compressed format of the video, specifically in terms of {\em motion displacements} encoding the motion between adjacent frames. As a result, the computational cost of our deformable attention is only $\mathcal{O}(S T^2 N)$. In our experiments we demonstrate that a small number of sample locations $N$ is sufficient to achieve strong performance (e.g., most of our results are obtained with $N=8$). This implies that, in practice, the complexity of our deformable attention is considerably lower than that of global attention, which is $\mathcal{O}(S^2 T^2)$ where $S >> N$. In addition to the efficiency benefits, our experiments show that our deformable attention achieves higher accuracy than global attention, attaining state-of-the-art results on four action classification benchmarks.

\section{Related Work}
\label{related_work}
\noindent\textbf{Image and Video Transformers:} The transformer architecture was originally introduced by Vaswani et al.~\cite{vaswani2017attention} as a model to capture long-range interactions between words in a sentence and was demonstrated to yield powerful representations for a variety of NLP tasks when trained with reconstruction objectives~\cite{devlin2018bert}. To extend transformers to the vision domain, Dosovitskiy et al.~\cite{dosovitskiy2020image} proposed the Vision Transformer (ViT), which applies a linear projection to tokenize non-overlapping patches of the image and then feeds the resulting tokens to a sequence of multi-headed self-attention blocks. Touvron et al.~\cite{touvron2020training} leveraged data augmentations and a student-teacher training scheme to increase the data efficiency of ViT. More recently, we have seen several attempts at extending ViT to the video domain~\cite{bertasius2021space, arnab2021vivit, akbari2021vatt, fan2021multiscale,patrick2021keeping}. TimeSformer~\cite{bertasius2021space}, ViViT~\cite{arnab2021vivit} and VideoSwin~\cite{liu2021video} treat time as an extra dimension and propose several temporal attention schemes to compare patches in different frames. MViT~\cite{fan2021multiscale} reduces significantly the computational cost of video transformers by means of a local pooling operation that reduces progressively the number of tokens while increasing the channel dimension. The attention scheme that is most closely related to our own is the ``trajectory attention'' of MotionFormer, which is introduced in concurrent work~\cite{patrick2021keeping}. This approach shares the same motivation as our strategy. However, it differs in two fundamental aspects. First, while our method leverages motion information available ``for free'' in the compressed video, trajectory attention requires the costly computation of a correlation volume expressing the probabilistic trajectories of each query patch over the clip. This causes trajectory attention to have quadratic cost $\mathcal{O}(S^2 T^2)$ and to be, in practice, twice as costly as global attention in terms of GFLOPs. Second, trajectory attention uses a hand-designed attention strategy giving more importance to tokens along the motion trajectory of the query patch. Conversely, our model uses the motion displacement (plus appearance information) as cues to predict ``where to look.'' However, the actual attention model is {\em learned} by optimizing the end-objective of video classification instead of being hand-defined. 



\noindent\textbf{Efficient Attention Schemes:} Several solutions have been proposed to reduce the quadratic cost of global self-attention.
One direction is to introduce manually defined local attention windows, and to restrict the attention operation to query-key pairs in the same local window. Longformer~\cite{beltagy2020longformer} and BigBird~\cite{zaheer2020big} implement this idea in the NLP domain, while Swin~\cite{liu2021video,liu2021swin} and CSwin Transformer~\cite{dong2021cswin} apply it to vision problems. Local windows, however, limit the span of the receptive field. To contrast this limitation, Longformer~\cite{beltagy2020longformer} and BigBird~\cite{zaheer2020big} allow a fraction of the tokens to use global attention, while Swin~\cite{liu2021video,liu2021swin} and CSwin Transformer~\cite{dong2021cswin} propose spatial merging and window shifting to increase the receptive field. 
In the NLP domain, Reformer~\cite{kitaev2020reformer} and Routing Transformer~\cite{roy2021efficient} use locality-sensitivity hashing and $k$-means to find the most relevant keys from all candidates. 
Deformable DETR~\cite{zhu2020deformable} follows a similar idea for the problem of object detection in still images. It uses a learned attention function to select the most salient patches to compare to the given query patch. The locations of the patches to attend are predicted from multi-scaled feature maps obtained with an independently-trained CNN. Our Deformable Video Transformer borrows the sampling strategy of Deformable DETR and extends it to the video domain by leveraging motion cues stored in compressed video to identify the most salient patches in frames different from the query patch. We also evaluate an optional multi-scale deformable attention to compare patches within the same frame. Differently from Deformable DETR, our multi-scale feature maps are obtained on the fly by means of a learnable convolutional layer attached to the attention block. This makes it possible to optimize the multiscale representation end-to-end together with the rest of the model with respect to the objective of video classification. Our experiments demonstrate that  multi-scale  attention provides synergistic information to that captured by deformable spatiotemporal attention, and that when these two attentions are combined together they lead to state-of-the-art results.



\noindent\textbf{Temporal Modeling in Video:} Analysis of the temporal information in the video is obviously critical for accurate action classification. Modern deep video architectures usually perform temporal analysis either by incorporating layers that can extract motion cues from raw RGB frames, such as 3D convolutions~\cite{JiEtAl:TPAMI2013, KarpathyEtAl:CVPR2014, DuEtAl:ICCV2015, DuEtAl:CVPR2018, XieEtAl:ECCV2018} and other learnable temporal operators~\cite{SunEtAl:CVPR2018, LeeEtAl:ECCV2018, hommos2018using, Lin_2019_ICCV, wang2020video}, or by feeding pre-extracted optical flow to the network~\cite{simonyan2014two,feichtenhofer2016convolutional, wang2016temporal}. More related to our own approach, a few works have proposed to leverage motion cues already available in compressed video (such as motion displacements stored by modern codecs, e.g., MPEG-4 and H.264)  for the purpose of action recognition~\cite{Kantorov_2014_CVPR, Zhang_2016_CVPR, wu2018compressed, korbar2019scsampler}. In this paper we also make use of these ``free'' motion cues in compressed video. However, instead of treating them as an additional input modality for action classification, we employ them to guide the spatiotemporal sampling of patches for self-attention.

\section{Technical Approach}
\label{sec:method}

We introduce our approach by first reviewing the general architecture of video transformers, which adapt the still-image framework of the Vision Transformer (ViT)~\cite{dosovitskiy2020image} to the video setting. We then describe our space-time deformable attention, which replaces the manually-defined attention mechanism of prior video transformers with an input-dependent, dynamic attention strategy that results in lower computational cost and higher accuracy.

\subsection{Background: Video Transformers}

Our method is designed to operate with any video transformer framework~\cite{bertasius2021space, neimark2021video, arnab2021vivit, fan2021multiscale, patrick2021keeping} and is not tied to a specific architecture.  A video transformer takes as input a video clip $X \in \mathbb{R}^{H \times W \times 3 \times T}$ consisting of $T$ RGB frames sampled from the original video. The video is converted into a sequence of $S \cdot T$ tokens ${\bf x}_s^t \in  \mathbb{R}^D$ for $s=1, \hdots, S$ and $t=1, \hdots, T$. The tokens ${\bf x}_s^t$ are obtained by decomposing each frame into $S$ patches which are then projected to a $D$-dimensional space through a learnable linear transformation. This tokenization can be implemented by linearly mapping the RGB patches of each frame~\cite{bertasius2021space,neimark2021video} or by projecting space-time cubes of adjacent frames~\cite{arnab2021vivit, fan2021multiscale,patrick2021keeping}. 
Separate learnable positional encodings ${\bf e}_s$ and ${\bf e}^t$ are then applied to the patch embeddings ${\bf x}_s^t$ for the spatial and the temporal dimension: ${\bf z}_s^t = {\bf x}_s^t + {\bf e}_s + {\bf e}^t$. 

The tokens are processed by a sequence of $L$ transformer layers applying multi-head attention (\text{MHA})~\cite{vaswani2017attention}, layer norm (\text{LN})~\cite{ba2016layer}, and multilayer perceptron (\text{MLP}) computation. Formally, layer $\ell$ transforms a token ${{\bf z}_s^t}^{(\ell)}$ into token ${{\bf z}_s^t}^{(\ell+1)}$ according to the following updates:
\begin{align}
&\hat{\bf z}_s^{t^{(\ell+1)}} = \text{MHA}\left(\text{LN}\left({\bf z}_s^{t^{(\ell)}}\right)\right) + {\bf z}_s^{t^{(\ell)}} \\
&{\bf z}_s^{t^{(\ell+1)}} = \text{MLP}\left(\text{LN}\left(\hat{\bf z}_s^{t^{(\ell+1)}}\right)\right) + \hat{\bf z}_s^{t^{(\ell+1)}}
\end{align}
where $\ell=1,\hdots,L$  and ${\bf z}_s^{t^{(0)}} = {\bf z}_s^t$.

While most video transformers maintain a constant space-time resolution $S \times T$ and fixed feature dimension $D$ across all layers, some works have proposed patch-pooling~\cite{fan2021multiscale} and patch-merging~\cite{liu2021video}, which progressively reduce the space-time sequence length.

In order to lighten the notation, we describe the self-attention operation in the simplified setting of a single head of attention and drop the layer superscripts. First, query-key-value vectors for each patch at space location $s$ and time instant $t$ are computed via linear projections:
\begin{eqnarray}
{\bf q}_s^t = W_q \text{LN}\left({\bf z}_s^t\right) \\
{\bf k}_s^t = W_k \text{LN}\left({\bf z}_s^t\right) \\
{\bf v}_s^t = W_v \text{LN}\left({\bf z}_s^t\right)
\end{eqnarray}
where $W_q$ $W_k$ and $W_v$ are matrices of learnable parameters. Next, the self-attention operation computes a weighted sum of the value vectors, where the combination coefficients are obtained by comparing the query to the keys. 

\noindent\textbf{Global space-time attention.} In global attention (also known as dense or ``joint space-time''~\cite{bertasius2021space} attention), each query patch is compared to all the other patches in the clip:
\begin{equation}
\hat{\bf z}^t_s = {\bf z}^t_s + \sum_{s'=1}^{S} \sum_{t'=1}^{T} \alpha_{s,s'}^{t,t'} {\bf v}_{s'}^{t'} 
\end{equation}
where the coefficients $\alpha_{s,s'}^{t,t'}$ are computed in terms of dot products between the query and the keys, and are normalized to sum up to 1:
\begin{equation}
\label{MHA}
\alpha_{s,s'}^{t,t'} = \frac{\exp{<{\bf q}_s^t, {\bf k}_{s'}^{t'}>}}{\sum_{s'', t''} \exp{<{\bf q}_s^t, {\bf k}_{s''}^{t''}>}}.
\end{equation}
As can be seen from~\cref{MHA}, the computational complexity of global space-time attention is $\mathcal{O}(S^2T^2)$. 

\noindent\textbf{Divided space-time attention.} In order to reduce the computational cost and to lighten the memory footprint, prior works~\cite{bertasius2021space, arnab2021vivit} have proposed to factorize the attention operation along the temporal
and spatial dimensions. {\bf Space} attention compares only patches within the same frame $t$:
\begin{equation}
\hat{\bf z}^t_s = {\bf z}^t_s + \sum_{s'=1}^{S} \alpha_{s,s'}^{t} {\bf v}_{s'}^{t}.
\end{equation}
Conversely, {\bf time} attention compares the query patch only to patches at the same spatial location $s$ in the other frames:
\begin{equation}
\hat{\bf z}^t_s = {\bf z}^t_s + \sum_{t'=1}^{T} \alpha_{s}^{t,t'} {\bf v}_{s}^{t'}.
\end{equation}
The computational costs of these factorized operations are dramatically lower than that of dense attention: $\mathcal{O}(S^2T)$ for space-only attention, and $\mathcal{O}(ST^2)$ for time-only attention. In order to allow the model to capture dependencies along both dimensions, space attention and temporal attention are combined within each layer either by composition~\cite{bertasius2021space} or by concatenation of their independent outputs~\cite{arnab2021vivit}.

The fundamental limitation of factorized self-attention is that the temporal attention module compares patches in different frames but at the {\em same} spatial location. However, due to camera and object motion, the same patch will capture disparate 3D scene regions at different times in the sequence. Temporal attention encodes appearance correlations between these different scene regions neglecting to take into account the dynamics of the scene. 


\subsection{Deformable Video Transformer}

In this subsection we introduce our Deformable Video Transformer, which uses motion cues to identify a sparse set of space-time location to attend for each query. Our motion cues are obtained at zero computational cost by leveraging motion displacements stored in the compressed format of the video and our learned attention module is directly optimized with respect to the end-objective of classification. 

\noindent\textbf{Deformable Space-Time Attention (D-ST-A).} Our proposed scheme resembles time-only attention but with two fundamental differences. The first is that, for each query ${\bf q}_s^t$, the method attends $N$ locations $\{s(n)\}_{n=1}^N$ within each frame $t'$, instead of a single location $s$. Thus, the computational cost becomes $\mathcal{O}(ST^2N)$. This is dramatically lower than the $\mathcal{O}(S^2T^2)$ complexity of dense space-time attention, since $N$ is more than two orders of magnitude smaller than the number of spatial patches $S$ (we use $N=8$ while $S=3,136$).
At the same time, attending multiple locations per frame allows the method to capture a bigger spatial context compared to vanilla time-only attention.

\begin{figure}
    \centering
    \includegraphics[width=1.0\linewidth]{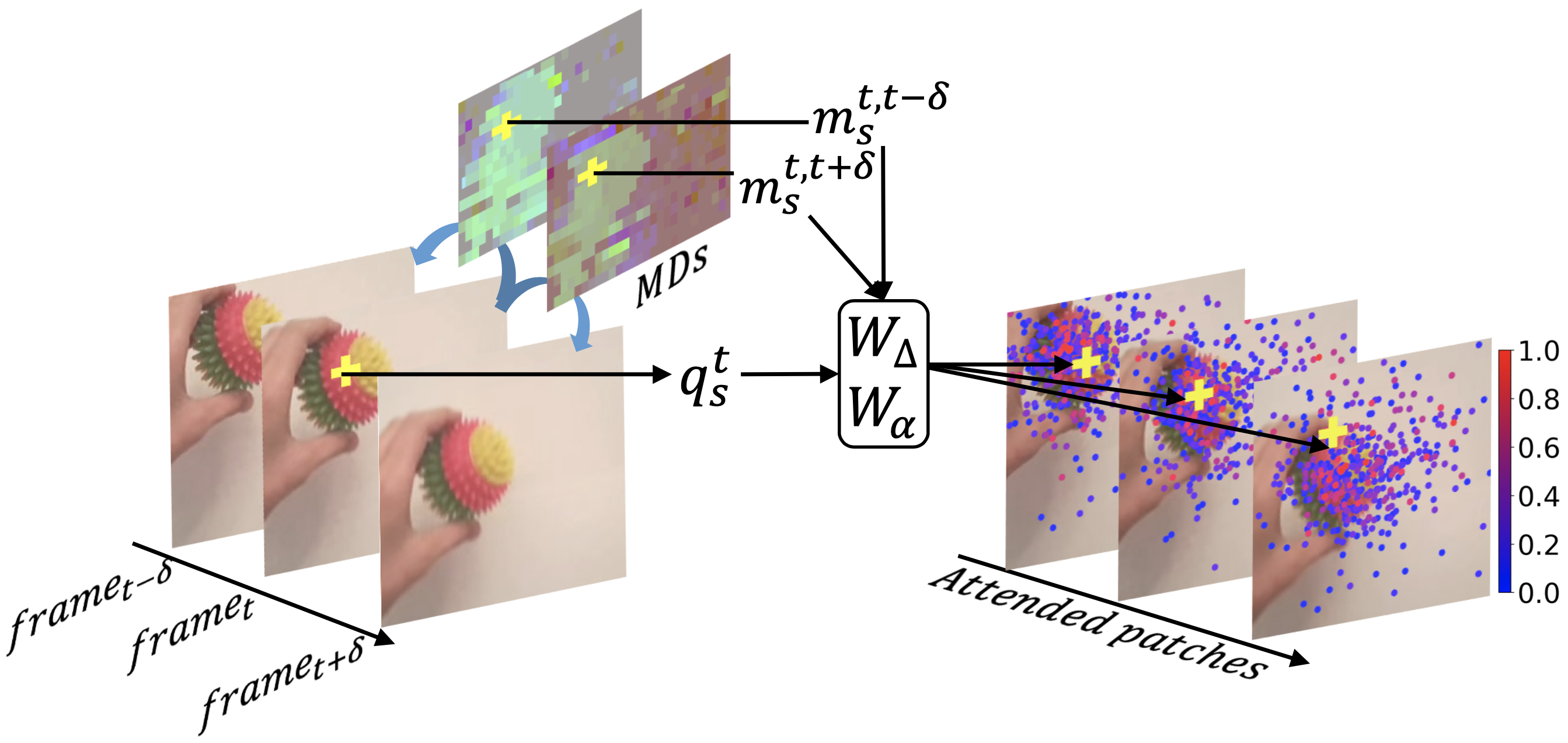}
    \caption{A visualization of D-ST-A. The yellow cross denotes a query patch in frame $t$. Each attention layer samples $N=8$ patches with associated attention weights from each frame based on motion displacements (MDs). Filled circles denote sampled patch centers and their color (from blue to red) indicates the value of the attention weights (from 0 to 1). Additional examples of attended patches are given in the supplementary material.}
    \label{Fig:D_attn}
\end{figure}

The second key difference is that these $N$ locations are input-conditioned, i.e., their positions are computed as a function of ${\bf q}_s^t$ {\em and} of the motion relating frame $t$ to $t'$. The intuition is that motion cues between the two frames provide salient information about where to look (i.e., which locations to attend) in frame $t'$. See Figure~\ref{Fig:D_attn} for a visualization of attended patches in frame $t-\delta$, $t$, and $t+\delta$, given a query in frame $t$. We refer the reader to the supplementary material for additional visualizations.

Formally, the update equation for the token at location $s$ in frame $t$ becomes:
\begin{equation}
\hat{\bf z}^t_s = {\bf z}^t_s + \sum_{t'=1}^{T} \sum_{n=1}^{N} \alpha_{s(n)}^{t,t'} {\bf v}_{s(n)}^{t'}. 
\label{eq:DSTA}
\end{equation}
The value vector ${\bf v}_{s(n)}^{t'}$ is computed via bilinear interpolation at 2D location ${\bf p}(s(n))$, which is obtained by adding an offset $\Delta_s^{t,t'}(n)$ to the 2D coordinates ${\bf p}(s)$ of patch $s$:
\begin{equation}
{\bf p}(s(n)) = {\bf p}(s) + \Delta_s^{t,t'}(n).
\end{equation}
The offset $\Delta_s^{t,t'}(n)$ is computed by means of a linear projection of the {\em appearance} embedding ${\bf z}^t_s$ and the {\em motion} embedding ${\bf m}^{t,t'}_s$:
\begin{equation}
\label{d_offset}
\Delta_s^{t,t'} = W_{\Delta} ({\bf q}_s^{t} +  {\bf m}_s^{t,t'})
\end{equation}
where $W_{\Delta} \in \mathbb{R}^{2N \times D}$ is a matrix of learnable parameters. 

\begin{figure}
    \centering
    \includegraphics[width=1.0\linewidth]{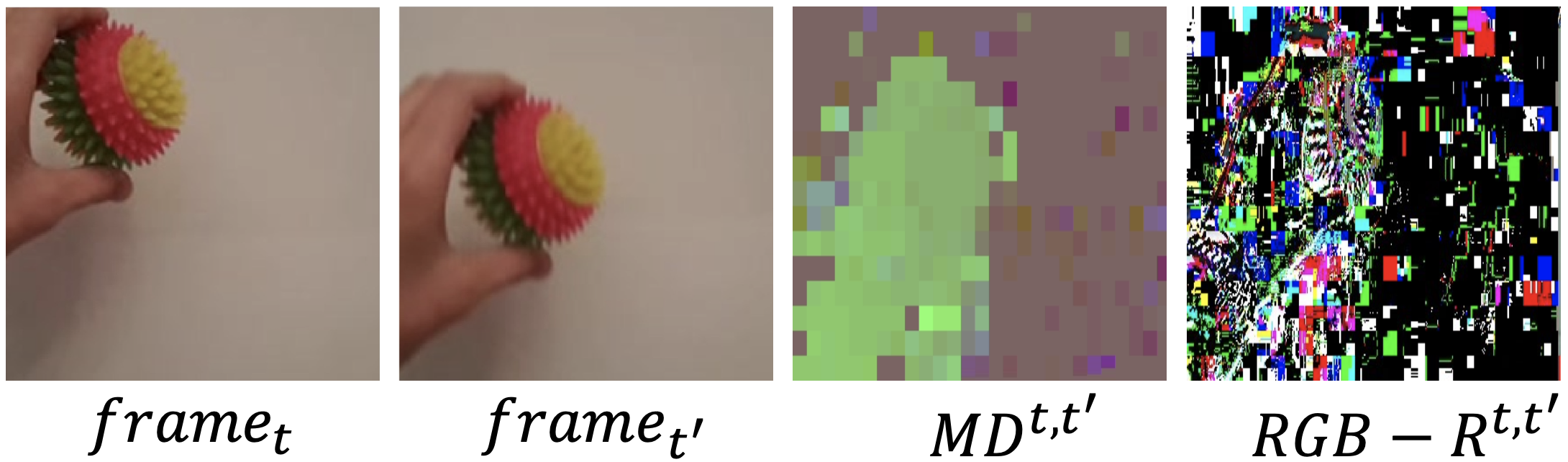}
    \caption{Example of motion displacements (MDs) and RGB residuals (RGB-Rs) between $frame_t$ and $frame_{t'}$.}
    \label{Fig:motion_example}
\end{figure}

To compute the motion embedding ${\bf m}_s^{t,t'}$ 
we utilize motion displacements and RGB residuals stored in the compressed video, which can both be easily accessed without any additional computation. Specifically, codecs such as MPEG-4 and H.264 represent compressed video in the form of sparsely stored I-frames, each followed by a sequence of 11 P-frames. An I-frame represents a frame in explicit RGB format at regular resolution, while the following P-frames encode implicitly the 11 subsequent RGB frames in terms of motion displacements (MD) and RGB residuals (RGB-R) which capture motion between adjacent frames and remaining RGB differences at low resolution, respectively. The RGB frames after an I-frame can be approximately reconstructed by rewarping the information stored in the I-frame according to the MD motion fields and by adding the remaining RGB-Rs. 
In this work, we accumulate the motion displacements and the RGB residuals between time step $t$ and $t'$ to represent the cumulative motion between these two frames from the information stored for the adjacent frames in-between. We then apply a patch decomposition and a learnable tokenization separately to the MDs and the RGB-Rs in order to obtain motion embeddings ${{\bf m}_s^{MD}}^{t,t'}, {{\bf m}_s^{RGBR}}^{t,t'} \in \mathbb{R}^D$. In the experiments we compare quantitatively these two different motion embeddings. Examples of motion displacements and RGB residuals are given in Figure~\ref{Fig:motion_example}.


Note that in our system the attention coefficients are linearly projected from the appearance and motion embeddings instead of being computed by pairwise comparisons of keys and values. Specifically, they are obtained by first computing the vector of coefficients $\hat{{\boldsymbol{\alpha}}}^{t,t'}_{s} \in \mathbb{R}^{N}$:
\begin{equation}
\label{d_attn}
\hat{{\boldsymbol{\alpha}}}^{t,t'}_{s} = W_{\alpha} ({\bf q}_s^{t} +  {\bf m}_s^{t,t'})
\end{equation}
with $W_{\alpha} \in \mathbb{R}^{N \times D}$ and, finally, by normalizing the $N$ coefficients per sample location:
\begin{equation}
{\alpha}^{t,t'}_{s(n)} = \frac{\exp{\hat{\alpha}^{t,t'}_{s(n)}}}{\sum_{t'=1}^T\sum_{n'=1}^N \exp{\hat{\alpha}^{t,t'}_{s(n')}}}.
\end{equation}
While the self-attention coefficients could also be obtained by means of a traditional dot-product between queries and keys, the linear projection has been shown to give equivalent results  at lower computational cost~\cite{zhu2020deformable, zhu2019deformable}.

We found empirically that motion embeddings $m_s^{t,t'}$ become noisy proxies of motion paths when the temporal distance $(t-t')$ between the two frames is large, due to error accumulation as well as visual nuisances such as camera movements, scenes changes and occlusions. 
To address this problem, we subdivide the input clip of $T$ frames into $B$ non-overlapping sub-clips, each containing $T'= T/B$ frames. We then restrict the accumulation of value vectors in Eq.~\ref{eq:DSTA} to include only terms from frames $t'$ that belong to the same sub-clip as the query.
Our experiments include an ablation showing the effects of different values of $B$ on the recognition accuracy.

\noindent\textbf{Deformable Multi-Scale Attention (D-MS-A).} Inspired by the strong benefits shown by the use of multi-scale feature maps in transformer architectures~\cite{zhu2020deformable, fan2021multiscale}, we integrate an optional deformable multi-scale attention (D-MS-A) in our design. While our deformable {\em space-time} attention is used to capture dependencies between different frames by means of motion embeddings, our deformable {\em multi-scale} attention is a form of ``space-only'' attention encoding correlations within the same frame as the query. 
For each frame $t$, this involves computing $F$ feature maps ${\bf z}_{s}^{(f),t}$ at different spatial resolutions $S^{(f)}$ for $f=1, \hdots, F$,
and then sampling $N'$ patches to attend at each scale. The multi-scale feature maps are obtained by attaching a learnable 3D convolutional layer 
to each block. Then, the token update equation is given by:
\begin{equation}
\hat{\bf z}^t_s = {\bf z}^t_s + \sum_{f=1}^{F} \sum_{n=1}^{N'} \alpha_{s^{(f)}(n)}^{t} {\bf v}_{s^{(f)}(n)}^{(f),t}
\end{equation}
where, as in the case of deformable space-time attention, the 2D coordinates ${\bf p}(s^{(f)}(n))$ of the $n$-th patch to attend at scale $f$ are obtained by adding an estimated offset to the 2D location of the query at that scale, ${\bf p}(s^{(f)})$. Thus, ${\bf p}(s^{(f)}(n)) = {\bf p}(s^{(f)}) + \Delta_s^{(f),t}(n)$. However, this time the offset is predicted from the features at scale $f$ for the {\em same} frame as the query:
\begin{equation}
\Delta_s^{(f),t} = W_{\Delta}^{(f)} {\bf q}_s^{(f),t}.
\end{equation}

\noindent\textbf{Attention Fusion.} In our experiments we present results using only deformable space-time attention (D-ST-A), only deformable multi-scale attention (D-MS-A), as well as a combination of the two (D-ST+MS-A). In the last case, our method feeds the two tokens, ${{\bf z}^{ST}}^t_s$ and ${{\bf z}^{MS}}^t_s$, computed independently by these two attention strategies to an {\em attention fusion} layer $u()$ which produces the token ${{\bf z}}^t_s$ to pass to the next block: ${{\bf z}}^t_s = u({{\bf z}^{ST}}^t_s, {{\bf z}^{MS}}^t_s)$. 
We present results for two forms of attention fusion, one based on a simple linear projection and the other based on the MLP-Mixer module~\cite{tolstikhin2021mlp}. The architecture details of these two modules are given in the supplementary material.

\noindent\textbf{Multiple Heads of Attention.} As customarily done in transformers, we use multiple heads of attention in each layer, allotting an equal number of dimensions to each head. The outputs of the individual heads are concatenated and projected through an MLP with a residual connection.

\section{Experiments}

\newif\ifablationtwo
\ablationtwofalse

\subsection{Implementation Details}
To demonstrate the generality of our Deformable Video Transformer (DVT), we apply it to two open-source video transformer architectures---MViT~\cite{fan2021multiscale} and TimeSformer~\cite{bertasius2021space}---where we substitute their traditional fixed attention strategies with our input-dependent deformable attention. 
We refer to these two architectures as MViT-B and ViT-B, respectively.

Unless otherwise noted, we use input clips of size $16\times224\times224$, sampled with temporal stride of 4. The clip-level classification is obtained by feeding the average-pooled features from the last layer to a fully connected layer. Video-level classification is performed by averaging the clip-level predictions computed from uniformly-spaced clips sampled from the video. 
The optimization recipe and the data augmentation strategy follow the implementation in~\cite{fan2021multiscale}. For our  deformable attention, we set $N=8$ and $B=4$ in both D-ST-A and D-MS-A by default, but we further ablate these hyperparameters in the experiments. In D-MS-A, to generate multi-scale feature maps we adopt a 3D convolutional layer with kernel size $3 \times 3 \times 3$, initial spatial stride of 8 (halved at each stage) and temporal stride of 1. 

\label{sec:exp}
\subsection{Datasets}
We evaluate our DVT on four standard video classification benchmarks: Kinetics-400~\cite{kay2017kinetics} (K400), Something-Something-V2~\cite{goyal2017something} (SSv2), EPIC-KITCHENS-100~\cite{damen2020rescaling} (EK100), and Diving-48~\cite{li2018resound} (D48). 

\subsection{Ablation Studies}
\label{sec:ablation}
We begin by studying how the accuracy varies as we modify the hyperparameters and the design choices in our DVT. Due to the high computational cost involved by these numerous ablations, here we limit these evaluations to the MViT architecture on the 
K400 and SSv2 benchmarks. The model on K400 is trained from random initialization (scratch). Since SSv2 is a much smaller dataset, we report results on it by finetuning the model pretrained on K400.

\noindent\textbf{Choice of motion cues.} In Table~\ref{motion_cue} we study the effectiveness of two different motion cues obtained from compressed video: RGB residuals (RGB-Rs) and motion displacements (MDs). We also consider a couple of simple cues directly computable from RGB values: ``Averaged Pooled RGB'' (Avg-P-RGB) averages the RGB values of the patches in query position $s$ between all frames between time $t$ and time $t'$; ``RGB Differences'' (RGB-D) uses the pixel-wise differences between the patches in query position $s$ between frame $t$ and frame $t'$ . We include also results obtained with TV-L1 Optical Flow~\cite{zach2007duality} (OF) since optical flow represents the ideal motion cue to use, neglecting cost considerations. Table~\ref{motion_cue} shows that Avg-P-RGB is the worst cue. RGB-D and RGB-R perform similarly and only slightly better than Avg-P-RGB. Conversely, MDs yield much higher accuracy than RGB-Rs, nearly on par with OF. This makes sense, since Avg-P-RGB, RGB-R and RGB-D capture color information which is only loosely related to the motion (e.g., as shown in Figure~\ref{Fig:motion_example}, RGB-R values tend to be higher in regions affected by motion, due to the need to fill-in the color of these regions to accurately reconstruct the RGB frames). Conversely, MDs encode true motion information between adjacent frames. Given the high computational cost of extracting OF, MD is an effective surrogate, achieving similar performance at no cost. Based on these results, we use MDs for all our subsequent experiments.
\begin{table}[]
\footnotesize
\centering
\begin{tabular}{|l|c|c|}
\hline
Motion Cue                      & K400 & SSv2 \\ \hline \hline
Averaged Pooled RGB (Avg-P-RGB) & 72.3\%   & 62.9\%  \\ 
RGB Residuals (RGB-R)      & 73.2\%     & 63.3\%     \\ 
RGB Differences (RGB-D) & 73.6\% & 63.5\%\\
Motion Displacements (MD) & 77.9\%     & 65.6\%    \\
Optical Flow (OF) & 78.2\% & 66.0\% \\ \hline
\end{tabular}
\caption{Accuracy of DVT with D-ST-A attention on K400 and SSv2 using different motion cues.}
\label{motion_cue}
\end{table}

\noindent\textbf{Attention and fusion blocks.} In Table~\ref{atten_fusion} we compare the performance of our two proposed attention schemes (D-ST-A and D-MS-A) as well as their combination (D-ST+MS-A) using either a linear projection (L) or MLP-Mixer~\cite{tolstikhin2021mlp} (M) as fusion block. We also include the results obtained with the original Multi Head Pooling Attention (MHPA) proposed in MViT~\cite{fan2021multiscale}.
\ifablationtwo
Here we include results using also the ViT-B architecture (bottom part of the Table). Since this architecture has been shown to require image-based pretraining to achieve strong accuracy~\cite{bertasius2021space}, we measure results on K400 using ImageNet-21K (IN-21K) pretraining, and those on SSv2 by pretraining on IN-21K and then K400.

The trend of results is quite similar under both architectures and on both datasets. D-ST-A gives consistently higher accuracy compared to D-MS-A. However, the two attention schemes clearly provide somewhat complementary information, as their combinations yield further accuracy gains. The best results are obtained with MLP-Mixer~\cite{tolstikhin2021mlp} as fusion block on both K400 and SSv2 for both architectures, although at a higher cost. 
\else
Compared to MHPA, DVT with simple D-ST-A has lower cost (57.4 vs 71 GFLOPs) and yields better accuracy on SSV2 ($65.6\%$ vs. $64.7\%$). On K400 is nearly on-par in terms of accuracy ($77.9\%$ vs $78.4\%$) while providing an efficiency gain. Merging D-MS-A with D-ST-A using our fusion module elevates substantially the accuracy.
The best results are obtained with MLP-Mixer~\cite{tolstikhin2021mlp} as fusion block on both K400 and SSv2, although at a higher cost. 
\fi
MLP-Mixer, specifically Channel-Mixer, projects the features into a higher dimensional space with a non-linear activation function, which allows better channel-wise communication compared to the linear projection. Based on these results, we use D-ST+MS-A with MLP-Mixer for the rest of the experiments.


\ifablationtwo

\begin{table}[]
\footnotesize
\centering
\begin{tabular}{|l|c|c|c|c|c|}
\hline
Attention  & ViT Architect.       & K400 & SSv2  &GFLOPs&Params   \\ \hline \hline
D-MS-A  & MViT-B     & 76.8\%       & 65.1\% &61.3 &32.4M\\ 
D-ST-A  & MViT-B     & 77.9\%       & 65.6\% &57.4 &31.9M\\ 
D-ST+MS-A (L) & MViT-B       & 78.1\%       & 66.8\%  & 81.9 &45.1M\\ 
D-ST+MS-A (M) & MViT-B       & \bf 79.0\%       & \bf 67.5\%  &128.3 &73.9M \\ \hline 
D-MS-A & ViT-B     & 78.3\%  & - & 160  &78.1\\ 
D-ST-A & ViT-B     & 79.5\%  & - & 147.5 & 76.3M\\ 
D-ST+MS-A (L) & ViT-B       & 80.1\% & - & 223.8 & 110.1M\\ 
D-ST+MS-A (M) & ViT-B      & \bf 80.5\% & - & 325 & 168.2M\\ \hline
\end{tabular}
\caption {Accuracy of DVT on K400 and SSv2 for our two proposed attention schemes (D-ST-A and D-MS-A) as well as their combination (D-ST+MS-A) using either a linear projection (L) or MLP-Mixer~\cite{tolstikhin2021mlp} (M) as fusion block. Top part shows results with the MViT-B architecture, bottom part reports results with ViT-B.}
\label{atten_fusion}
\end{table}

\else

\begin{table}[]
\footnotesize
\centering
\begin{tabular}{|l|c|c|c|c|}
\hline
Attention  & K400 & SSv2  &GFLOPs & Parameters  \\ \hline \hline
MHPA~\cite{fan2021multiscale}  & 78.4\%       & 64.7\%  &70.5 &36.6M  \\ \hline
D-MS-A  &  76.8\%     & 65.1\% &61.3 &32.4M \\ 
D-ST-A  &  77.9\%     & 65.6\% &57.4 & 31.9M \\ 
D-ST+MS-A (L) &  78.1\%       & 66.8\%  & 81.9 & 45.1M \\ 
D-ST+MS-A (M) &  \bf 79.0\%       & \bf 67.5\%  &128.3 & 73.9M \\ \hline 
\end{tabular}
\caption {Accuracy of DVT on K400 and SSv2 for our two proposed attention schemes (D-ST-A and D-MS-A) as well as their combination (D-ST+MS-A) using either a linear projection (L) or MLP-Mixer~\cite{tolstikhin2021mlp} (M) as fusion block. The architecture is MViT-B.}
\label{atten_fusion}
\end{table}

\fi

\begin{figure}[]
\centering
\begin{subfigure}{.235\textwidth}
  \includegraphics[width=1.0\linewidth]{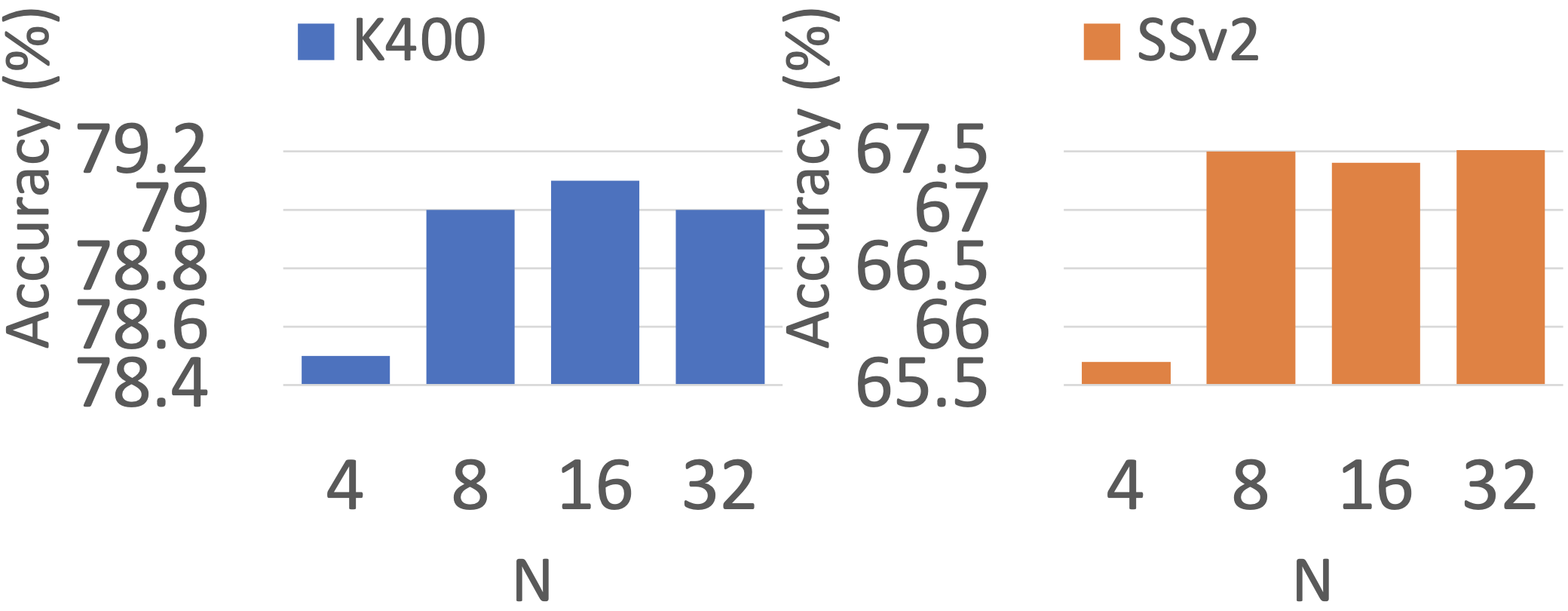}
  \caption{}
  \label{fig:N_point}
\end{subfigure}
\begin{subfigure}{.235\textwidth}
  \includegraphics[width=1.0\linewidth]{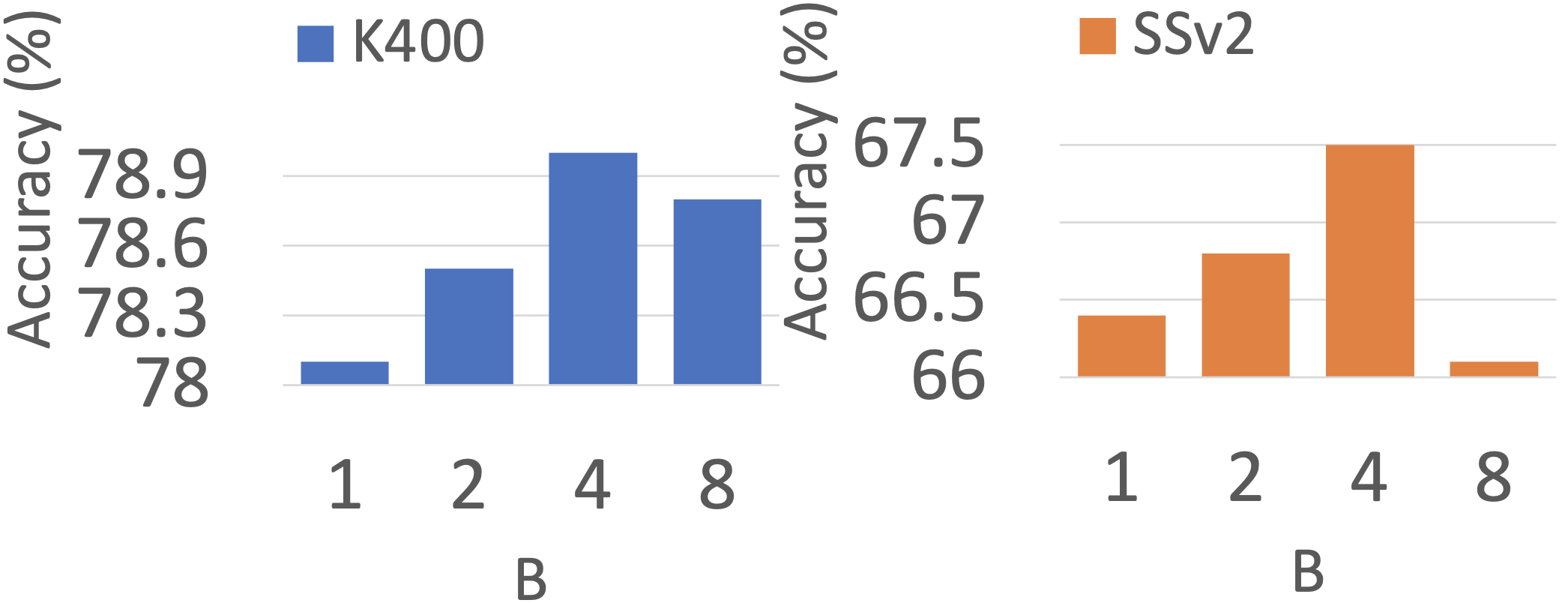}
  \caption{}
  \label{fig:N_window}
\end{subfigure}
\caption{Accuracy of DVT on K400 and SSv2 when varying (a) the number of attended patches in each frame, $N$, or (b) the number of subclips for interframe comparison, $B$. The attention strategy is D-ST+MS-A (M) and the architecture is MViT-B.}
\end{figure}

\noindent\textbf{Number of patches in the deformable attention.} 
In Figure~\ref{fig:N_point} we present the accuracy of our DVT model for a varying number of attended patches, $N$, in each frame, using D-ST+MS-A as attention. We see that while accuracy increases when moving from $N=4$ to $N=8$, the accuracy tends to level off after $N=8$. This confirms that the frames in the video contain highly redundant information and that it is sufficient to attend a small number of patches to achieve strong accuracy. Since this hyperparameter affects the computational cost of our model, we set $N=8$ for the subsequent studies as this represents a good trade-off in terms of accuracy versus efficiency.

\noindent\textbf{Number of sub-clips in D-ST-A.} In Figure~\ref{fig:N_window} we report accuracy achieved when varying the number of attention sub-clips $B$ in each clip. Each sub-clip contains $T'=T/B$ frames where $T$ is the number of frames in the clip (here we use $T=16$). D-ST-A compares only frames belonging to the same sub-clip. We see that the best accuracy is achieved for $B=4$, which entails comparing $T'=4$ frames within each sub-clip. Using smaller values for $B$ yields lower accuracy as motion displacements become unreliable on longer sub-clips. Conversely, using $B=8$ limits the ability of D-ST-A to leverage temporal information, which is particularly detrimental on SSv2. 

\noindent\textbf{Per-Class Comparison.} In Table~\ref{per_class}, we show the five K400 classes that receive the largest gain/degradation in accuracy from the use of D-ST-A compared to the same model (MViT) using MHPA. It can be seen that D-ST-A delivers much bigger gains for the top-5 classes, compared to the small drops in accuracy on the five classes where D-ST-A is most detrimental.
\begin{table}[!ht]
\footnotesize
\centering\vspace{-0.2cm}
\begin{tabular}{|l|c|l|c|}
\hline
Improved Classes & Change & Degraded Classes & Change\\ \hline \hline
Eating Doughnuts & +40.8\% & Busking & -8.0\%  \\ \hline
Using Controller & +40.7\%  & Dining & -6.1\% \\ \hline
Drinking Shots & +37.5\%  &Diving Cliff &-6.1\% \\ \hline
Passing Football &+32.7\% & Pumping Gas &-4.2\% \\\hline
Peeling Potatoes &+30.0\% &Motorcycling &-4.1\% \\\hline
\end{tabular}\vspace{-0.0cm}
\caption{K400 classes receiving the largest gain/degradation in accuracy when inserting D-ST-A in the MViT architecture.}
\label{per_class}
\end{table}\vspace{-0.0cm}

\subsection{Comparison to the state-of-the-art}

In Tables~\ref{k400},~\ref{SSv2},~\ref{EK-100},~\ref{DV-48} we compare our method against the state-of-the-art on K400, SSv2, EK100, and D48. We note that the accuracy of an action recognition model depends strongly on (i) the architecture, (ii) the pretraining dataset, (iii) the clip size. In order to make the comparison with prior works as fair as possible, we evaluated
our attention model under the many choices for (i, ii, iii) from the literature. To facilitate the interpretation of results, we split each Table into blocks, so that entries within the same block are results obtained with comparable (i, ii, iii) setups and differing with respect to the attention scheme (our contribution).
The rows in bold are the results of our method.

Based on the results in Tables~\ref{k400},~\ref{SSv2},~\ref{EK-100},~\ref{DV-48}, we can draw several observations. First, for the case of the MViT-B architecture, we can compare results obtained with our deformable attention against those achieved by the original Multi Head Pooling Attention (MHPA) proposed in MViT~\cite{fan2021multiscale}. Compared to MHPA, our D-ST+MS-A provides a Top-1 accuracy gain of $+0.6\%$ on K400 ($79.0\%$ vs $78.4\%$ in the learning from scratch setting) and of $+2.8\%$ on SSv2 ($67.5\%$ vs $64.7\%$ when both using K400 pretraining). SSv2 is a benchmark designed to require effective temporal modeling. The larger gain on SSv2 compared to K400 suggests that our model beneficially leverages the motion cues to capture relevant temporal dependencies.  In the case of the ViT-B architecture, we can compare our deformable attention against Divided-ST attention of TimeSformer~\cite{bertasius2021space} and Trajectory attention of MotionFormer~\cite{patrick2021keeping}, since both these works reported results using this architecture. Compared to Trajectory attention, our D-ST+MS-A achieves an accuracy gain of $0.8-0.9\%$ on K400 ($80.5\%$ vs $79.7\%$ for $16 \times 224^2$ clips and $82.0\%$ vs $81.1\%$ for $16 \times 336^2$ clips). Furthermore, our DVT is 7 times more efficient than MotionFormer in terms of GFLOPs. Our D-ST+MS-A improves over Trajectory attention by $+0.5-0.8\%$ on SSv2 ($67.0\%$ vs $66.5\%$ for $16 \times 224^2$ and $67.9\%$ vs $67.1\%$ for $16 \times 336^2$ clips) and by $+0.7\%$ on Action prediction ($43.8\%$ vs $43.1\%$) $+1.2\%$ on Verb prediction in EK100 ($68.2.8\%$ vs $67.0\%$) when both models use the same clip size. Compared to Divided-ST, our D-ST+MS-A produces an accuracy gain of $+3.8\%$ on K400 ($84.5\%$ vs $79.7\%$) when both use a clip size of $16 \times 448^2$. On D48, which is another benchmark designed to assess temporal modeling abilities, D-ST+MS-A yields an accuracy gain of $+6.6\%$ over Divided-ST ($86.6\%$ vs $78.0\%$) given the same input resolution. Finally, we also include results of DVT with the MViT-B architecture using longer clips ($32\times224^2$) which allow our method to achieve further improvements over the state-of-the-art on all four benchmarks. 

\begin{table*}[!ht]
\centering
\scriptsize
\begin{tabular}{|l|c|c|c|c|c|c|c|}
\hline
Method                                                          & Attention  & ViT Architecture & Pretraining & Clip Size       & Top-1                            & Top-5                            & GFLOPs x views \\ \hline \hline
SlowFast~\cite{feichtenhofer2019slowfast} & -          & -                & -           & $8\times224^2$  & 79.8\%                           & 93.9\%                           & $234\times3\times10$      \\
X3D~\cite{feichtenhofer2020x3d}           & -          & -                & -           & $16\times312^2$ & 79.1\%                           & 93.9\%                           & $48\times3\times10$      \\ \hline \hline

Tformer~\cite{bertasius2021space}         & Divided-ST & ViT-B            & IN-21K      & $8\times224^2$  & 78.0\%                           & 93.7\%                           & $197\times3\times1$        \\
Mformer~\cite{patrick2021keeping}         & Trajectory & ViT-B            & IN-21K      & $16\times224^2$ & 79.7\%                           & 94.2\%                           & $397\times3\times10$     \\
\bf DVT (ours)                                                      & \bf D-ST+MS-A  &\bf ViT-B            & \bf IN-21K      &  $\bf 16\times224^2$ & \bf 80.5\%                           & \bf 94.7\%                           &$ \bf 325\times1\times5 $       \\
Mformer~\cite{patrick2021keeping}         & Trajectory & ViT-B            & IN-21K      & $16\times336^2$ & 81.1\%                           & 95.2\%        & $959\times3\times10$     \\
\bf DVT (ours)                                                      & \bf D-ST+MS-A  &\bf ViT-B            & \bf IN-21K      &  $\bf 16\times336^2$ & \bf 82.0\%                           & \bf 95.3\%                           &$ \bf 731\times1\times5 $       \\
Tformer~\cite{bertasius2021space}         & Divided-ST & ViT-B            & IN-21K      & $16\times448^2$  & 79.7\%                           & 94.4\%                           & $1703\times3\times1$        \\

\bf DVT (ours)                                                    & \bf D-ST+MS-A &\bf ViT-B           & \bf IN-21K      &  $\bf 16\times448^2$ & \bf 84.5\%                          & \bf 96.7\%                           &$\bf 1300\times1\times5 $   \\

Mformer~\cite{patrick2021keeping}          & Trajectory & ViT-B            & IN-21K      & $32\times224^2$ & 80.2\%                           & 94.8\%                           & $1185\times3\times10$    \\
Tformer~\cite{bertasius2021space}        & Divided-ST & ViT-B            & IN-21K      & $96\times224^2$ & 80.7\%                           & 94.7\%                           & $2380\times3\times1$       \\ \hline 
ViViT~\cite{arnab2021vivit}               & Joint ST   & ViT-L            & IN-21K      & $16\times320^2$ & 81.3\%                           & 94.7\%                           & $3992\times3\times4$       \\ \hline\hline
MViT~\cite{fan2021multiscale}             & MHPA       & MViT-B           & -           & $16\times224^2$ & 78.4\%                           & 93.5\%                           & $71\times1\times5$       \\
\bf DVT (ours)                                                      &\bf D-ST+MS-A  &\bf MViT-B           & \bf-           & $\bf 16\times224^2$ &\bf 79.0\%                           & \bf 93.8\%    &$\bf 128\times1\times5$      \\ \hline 
\bf DVT (ours)                                                      &\bf  D-ST+MS-A  & \bf MViT-B           & \bf IN-1K       & $\bf 16\times224^2$ &\bf  80.8\%                           &\bf 95.0\%                           &$\bf 128\times1\times5$      \\ 
\bf DVT (ours)                                                      & \bf D-ST+MS-A  & \bf MViT-B           &\bf IN-21K      &  $\bf 16\times224^2$ & \bf 81.5\%                           & \bf 95.2\%                           &$ \bf 128\times1\times5$      \\ 
\bf DVT (ours)                                                     &\bf D-ST+MS-A  & \bf MViT-B          &\bf IN-21K      &  $\bf 32\times224^2$ & \bf 84.7\%                         & \bf 96.2\%                           &$\bf 259\times1\times5$  \\\hline 
\end{tabular}\vspace{-.3cm}
\caption{Video-level classification accuracy on Kinetics-400.\vspace{-.0cm}}
\label{k400}
\end{table*}

\begin{table*}[!ht]
\centering
\scriptsize
\begin{tabular}{|l|c|c|c|c|c|c|}
\hline
Method                                                   & Attention          & ViT Architect.  & Pretraining          & Clip Size                & Top-1                                 & Top-5                                \\ \hline \hline
MSNet~\cite{kwon2020motionsqueeze} & -                  & -               & IN-1K                & $16\times224^2$          & 64.7\%                                & 89.4\%                                   \\
bLVNet~\cite{fan2019more}          & -                  & -               & IN-1K                & $32\times224^2$          & 65.2\%                                & 90.3\%                                         \\ \hline \hline

Tformer~\cite{bertasius2021space}  & Divided-ST         & ViT-B           & IN-1K                & $8\times224^2$           & 59.5\%                                & 74.9\%                                                \\
\textbf{DVT (ours)}                                     & \textbf{D-ST+MS-A} & \textbf{ViT-B}  &\textbf{IN-1K} & $\bf 8\times224^2$ & \bf 64.8\%   & \bf 89.5\%      \\
Tformer~\cite{bertasius2021space}         & Divided-ST & ViT-B            & IN-1K      & $16\times448^2$  & 62.2\%                           & 78.0\%                                   \\
Tformer~\cite{bertasius2021space}  & Divided-ST         & ViT-B           & IN-1K                & $96\times224^2$          & 62.4\%                                & 81.0\%                                            \\ \hline
Mformer~\cite{patrick2021keeping}  & Trajectory         & ViT-B           & IN-21K+K400          & $16\times224^2$          & 66.5\%                                & 90.1\%                                       \\
\textbf{DVT (ours)}                                      & \textbf{D-ST+MS-A} & \textbf{ViT-B}  & \textbf{IN-21K+K400} & $\bf 16\times224^2$ & \bf 67.0\%   & \bf 90.5\%      \\
Mformer~\cite{patrick2021keeping}  & Trajectory         & ViT-B           & IN-21K+K400          & $16\times336^2$          & 67.1\%                                & 90.6\%                                        \\
\textbf{DVT (ours)}                                     & \textbf{D-ST+MS-A} & \textbf{ViT-B}  & \textbf{IN-21K+K400} & $\bf 16\times336^2$ & \bf 67.9\%   & \bf 90.8\%     \\

Mformer~\cite{patrick2021keeping}  & Trajectory         & ViT-B           & IN-21K+K400          & $32\times224^2$          & 68.1\%                                & 91.2\%                                        \\ \hline
ViViT~\cite{arnab2021vivit}        & Joint-ST           & ViT-L           & IN-21K               & $16\times320^2$          & 65.4\%                                & 89.8\%                                      \\ \hline\hline
MViT~\cite{fan2021multiscale}      & MHPA               & MViT-B          & K400                 & $16\times224^2$          & 64.7\%                                & 89.2\%                                     \\
\textbf{DVT (ours)}                                      & \textbf{D-ST+MS-A} & \textbf{MViT-B} & \textbf{K400}        & $\bf 16\times224^2$ & \textbf{67.5\%}                       & \textbf{90.8\%}                       \\ \hline
\textbf{DVT (ours)}                                      & \textbf{D-ST+MS-A} & \textbf{MViT-B} & \textbf{IN-1K}       & $\bf 16\times224^2$ & \textbf{67.4\%}                       & \textbf{90.6\%}                        \\ 
\textbf{DVT (ours)}                                      & \textbf{D-ST+MS-A} & \textbf{MViT-B} & \textbf{IN-21K}      & $\bf 16\times224^2$ & \bf 67.8\% & \bf 90.6\% \\ 

\textbf{DVT (ours)}                                      & \textbf{D-ST+MS-A} & \textbf{MViT-B} & \textbf{IN-1K+K400}  & $\bf 16\times224^2$ & \textbf{67.8\%}                       & \textbf{90.6\%}                       \\ 
\textbf{DVT (ours)}                                      & \textbf{D-ST+MS-A} & \textbf{MViT-B} & \textbf{IN-21K+K400} &$\bf 16\times224^2$ & \bf 68.0\% & \bf 91.0\%   \\
\textbf{DVT (ours)}                                     & \textbf{D-ST+MS-A} & \textbf{MViT-B} & \textbf{IN-21K+K400} &$\bf 32\times224^2$ & \bf 68.5\% & \bf 91.0\%  \\\hline 
\end{tabular}\vspace{-.3cm}
\caption{Video-level classification accuracy on Something-Something-V2.\vspace{-.0cm}}
\label{SSv2}
\end{table*}

\begin{table*}[!ht]
\centering
\scriptsize
\begin{tabular}{|l|c|c|c|c|c|c|c|}
\hline
Method                                                  & Attention          & ViT Architecture & Pretraining          & Clip Size                & A                                    & V                                    & N                                    \\ \hline \hline
TSM~\cite{lin2019tsm} & - & - & IN-1K &$16\times224^2$ &38.3\% &67.9\% &49.0\% \\
\hline\hline
Mformer~\cite{patrick2021keeping} & Trajectory         & ViT-B            & IN-21K+K400          & $16\times224^2$          & 43.1\%                               & 66.7\%                               & 56.5\%                               \\
\textbf{DVT (ours)}                                     & \textbf{D-ST+MS-A} & \textbf{ViT-B}   & \textbf{IN-21K+K400} & $\bf16\times224^2$ & \bf 43.8\% & \bf 67.5\% & \bf 56.5\%  \\
Mformer~\cite{patrick2021keeping} & Trajectory         & ViT-B            & IN-21K+K400          & $16\times336^2$          & 44.5\%                               & 67.0\%                               & 58.5\%            \\
\textbf{DVT (ours)}                                    & \textbf{D-ST+MS-A} & \textbf{ViT-B}   & \textbf{IN-21K+K400} & $\bf16\times336^2$ & \bf 45.0\% & \bf 68.2\% & \bf 58.7\%  \\
Mformer~\cite{patrick2021keeping} & Trajectory         & ViT-B            & IN-21K+K400          & $32\times224^2$          & 44.1\%                               & 67.1\%                               & 57.6\%                               \\ \hline
ViViT~\cite{arnab2021vivit}       & Joint ST           & ViT-L            & IN-21K+K400          & $16\times224^2$          & 44.0\%                               & 66.4\%                               & 56.5\%                               \\ \hline\hline
\textbf{DVT (ours)}                                     & \textbf{D-ST+MS-A} & \textbf{MViT-B}  & \textbf{IN-1K}       & $\bf16\times224^2$ & \textbf{42.3\%}                      & \textbf{68.4\%}                      & \textbf{54.4\%}                      \\ 
\textbf{DVT (ours)}                                     & \textbf{D-ST+MS-A} & \textbf{MViT-B}  & \textbf{IN-1K+K400}  &$\bf 16\times224^2$ & \textbf{43.3\%}                      & \textbf{68.9\%}                      & \textbf{55.6\%}                      \\
\textbf{DVT (ours)}                                     & \textbf{D-ST+MS-A} & \textbf{MViT-B}  & \textbf{IN-21K+K400} & $\bf 16\times224^2$ & \textbf{44.2\%}                      & \textbf{69.7\%}                      & \textbf{56.6\%}                      \\
\textbf{DVT (ours)}                                    & \textbf{D-ST+MS-A} & \textbf{MViT-B}  & \textbf{IN-21K+K400} & $\bf 32\times224^2$ & \textbf{45.8\%}                     & \textbf{69.9\%}                      &\textbf{58.7\%}                      \\\hline 
\end{tabular}\vspace{-.3cm}
\caption{Accuracy of Action (A), Verb (V) and Noun (N) classification achieved by different models on EPIC-KITCHENS.\vspace{-.0cm}}
\label{EK-100}
\end{table*}

\begin{table*}[!ht]
\centering
\scriptsize
\begin{tabular}{|l|c|c|c|c|c|}
\hline
Method                                                          & Attention          & ViT Architecture & Pretraining    & Clip Size                  & Top-1                              \\ \hline \hline
SlowFast~\cite{feichtenhofer2019slowfast} & -                  & -                & IN-1K          & $16 \times 224^2$          & 77.6\%                             \\ \hline \hline

Tformer~\cite{bertasius2021space}         & Divided-ST         & ViT-B            & IN-1K          & $8 \times 224^2$           & 74.9\%                             \\
\textbf{DVT (ours)}                                             & \textbf{D-ST+MS-A} & \textbf{ViT-B}   & \textbf{IN-1K} & $\bf 16 \times 224^2$ & \bf 83.5\% \\
Tformer~\cite{bertasius2021space}         & Divided-ST         & ViT-B            & IN-1K          & $16 \times 448^2$          & 78.0\%                             \\
\textbf{DVT (ours)}                                            & \textbf{D-ST+MS-A} & \textbf{ViT-B}   & \textbf{IN-1K} & $\bf 16 \times 448^2$ & \bf 86.6\% \\
Tformer~\cite{bertasius2021space}         & Divided-ST         & ViT-B            & IN-1K          & $96 \times 224^2$          & 81.0\%                             \\ \hline\hline
\textbf{DVT (ours)}                                             & \textbf{D-ST+MS-A} & \textbf{MViT-B}  & \textbf{IN-1K} & $\bf 16 \times 224^2$ & \textbf{86.0\%}                    \\ 
\textbf{DVT (ours)}                                             & \textbf{D-ST+MS-A} & \textbf{MViT-B}  & \textbf{IN-1K} & $\bf 32 \times 224^2$ & \textbf{86.9\%}                    \\\hline 
\end{tabular}\vspace{-.3cm}
\caption{Video-level classification accuracy on the Diving-48 dataset.\vspace{-.0cm}}
\label{DV-48}
\end{table*}
\section{Conclusion}
We have introduced a novel self-attention strategy for video classification. Unlike prior schemes that compare patches at predetermined locations, our method leverages motion cues to identify the most salient patches to compare to each query patch. The motion cues are obtained at zero cost from compressed video. This allows our method to achieve higher accuracy and lower computational cost compared to fixed schemes of attention. 
Currently, our approach limits the temporal span of attention to prevent motion errors to degrade the results. In the future, we are interested in exploring alternative motion cues that can be computed efficiently~\cite{SunEtAl:CVPR2018, LeeEtAl:ECCV2018, hommos2018using, Lin_2019_ICCV} and that may provide more reliable estimates of motion over longer temporal extents.


{\small
\bibliographystyle{ieee_fullname}
\bibliography{egbib}
}

\newpage 

\appendix
\section{Fusion Module: Implementation Details}

\begin{figure}[!ht]
\centering
   \begin{subfigure}{0.35\textwidth}
      \includegraphics[width=1.0\linewidth]{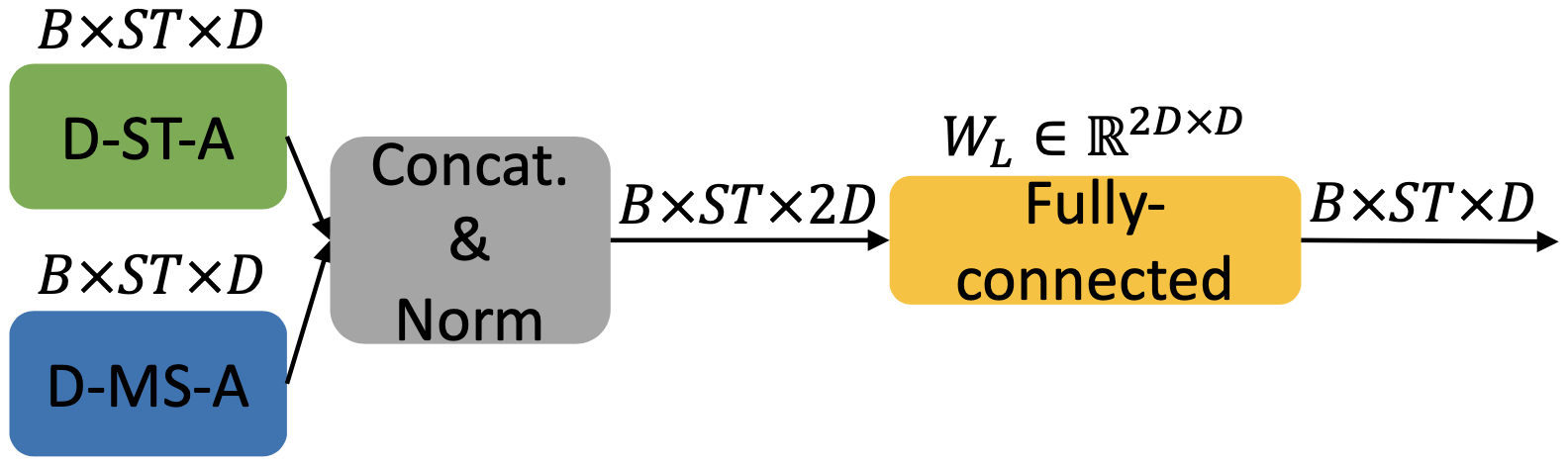}
      \caption{Linear Projection}
      \label{Fig:LP}
    \end{subfigure}
    \begin{subfigure}{0.49\textwidth}
      \includegraphics[width=1.0\linewidth]{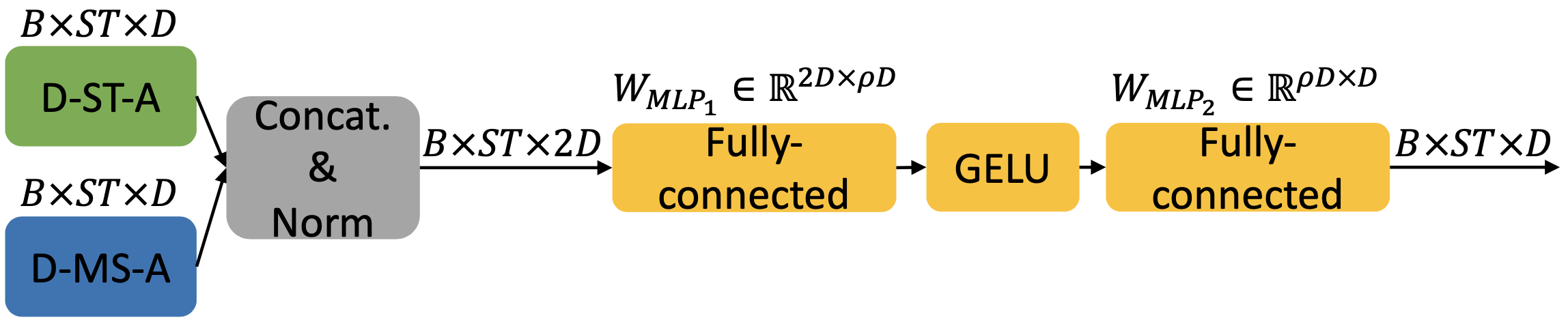}
      \caption{MLP-Mixer}
      \label{Fig:MLP}
    \end{subfigure}
    \caption{ 
    Illustration of the two fusion module variants considered for our DVT.\vspace{-.4cm}}
    \label{Fig:fusion}
\end{figure}

The fusion module combines in each layer the features obtained from our two deformable attention schemes (D-ST-A and D-MS-A). In this section we provide additional implementation details about our two fusion module variants: Linear Projection and MLP-Mixer.  Linear Projection (Figure~\ref{Fig:LP}) is a fully-connected layer which simply projects the concatenated feature from $2D$ dimensions to $D$ dimensions. Instead,   MLP-Mixer (Figure~\ref{Fig:MLP}) leverages the Channel-Mixer scheme~\cite{tolstikhin2021mlp}. Compared to Linear Projection, the first fully-connected layer of MLP-Mixer projects the feature into a higher dimensional space with a non-linear activation function and then projects it back to $D$ dimensions through another fully-connected layer. The hyperparameter $\rho$ controls the dimension of the higher dimensional space. In our experiments we set $\rho=4$. 

\begin{table}[!ht]
\footnotesize
\centering
\begin{tabular}{|l|c|c|c|c|}
\hline
Attention  & ViT Architect.       & K400 & SSv2    \\ \hline \hline
D-MS-A & ViT-B     & 78.3\%  & 64.1   \\ 
D-ST-A & ViT-B     & 79.5\%  & 65.3 \\ 
D-ST+MS-A (L) & ViT-B       & 80.1\% & 66.5\\ 
D-ST+MS-A (M) & ViT-B      & \bf 80.5\% & \bf 67.0  \\ \hline 
\end{tabular}
\caption {Accuracy of DVT on K400 and SSv2 for our two proposed attention schemes (D-ST-A and D-MS-A) as well as their combination (D-ST+MS-A) using either a linear projection (L) or MLP-Mixer~\cite{tolstikhin2021mlp} (M) as fusion block. The architecture is ViT-B. \vspace{-.4cm}}
\label{atten_fusion2}
\end{table} 

\section{Ablation on Attention and Fusion Modules}

In Section~\ref{sec:ablation} and Table~\ref{atten_fusion} of our main paper we presented results of an ablation on our two proposed attention schemes (D-ST-A and D-MS-A) and two fusion block variants (Linear Projection and MLP-Mixer) using the MViT-B architecture. In Table~\ref{atten_fusion2} of this this supplementary material we present results of the same ablation but for the case of the ViT-B architecture. We use an input clip of size $16\times224^2$ and we measure results on K400 using ImageNet-21K (IN-21K) pretraining, and those on SSv2 by pretraining on IN-21K and then K400. 
We can observe that the trend of results in Table~\ref{atten_fusion2} matches that already exhibited in Table~\ref{atten_fusion} of our main paper for the MViT-B architecture. D-ST-A gives consistently higher accuracy compared to D-MS-A. Furthermore, the fusion of the two attentions yields further accuracy gains, with the best results obtained with MLP-Mixer~\cite{tolstikhin2021mlp} on both K400 and SSv2. 

\begin{figure*}[!ht]
\centering
\begin{subfigure}{1.0\textwidth}
  \includegraphics[width=1.0\linewidth]{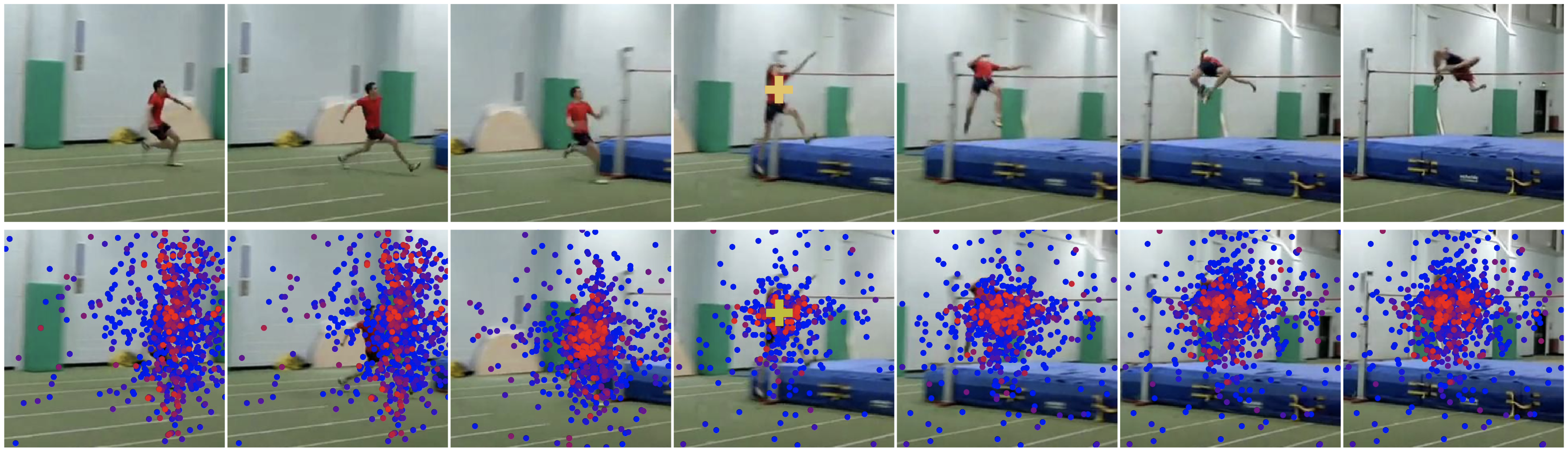}\vspace{1cm}
\end{subfigure}
\begin{subfigure}{1.0\textwidth}
  \includegraphics[width=1.0\linewidth]{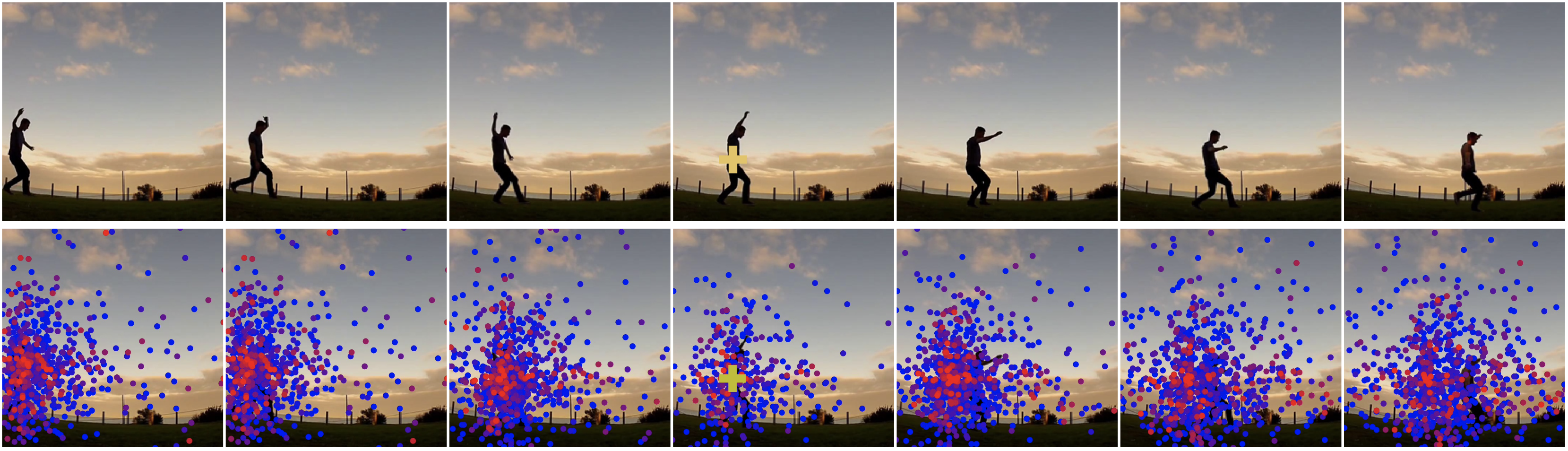}\vspace{1cm}
\end{subfigure}
\begin{subfigure}{1.0\textwidth}
  \includegraphics[width=1.0\linewidth]{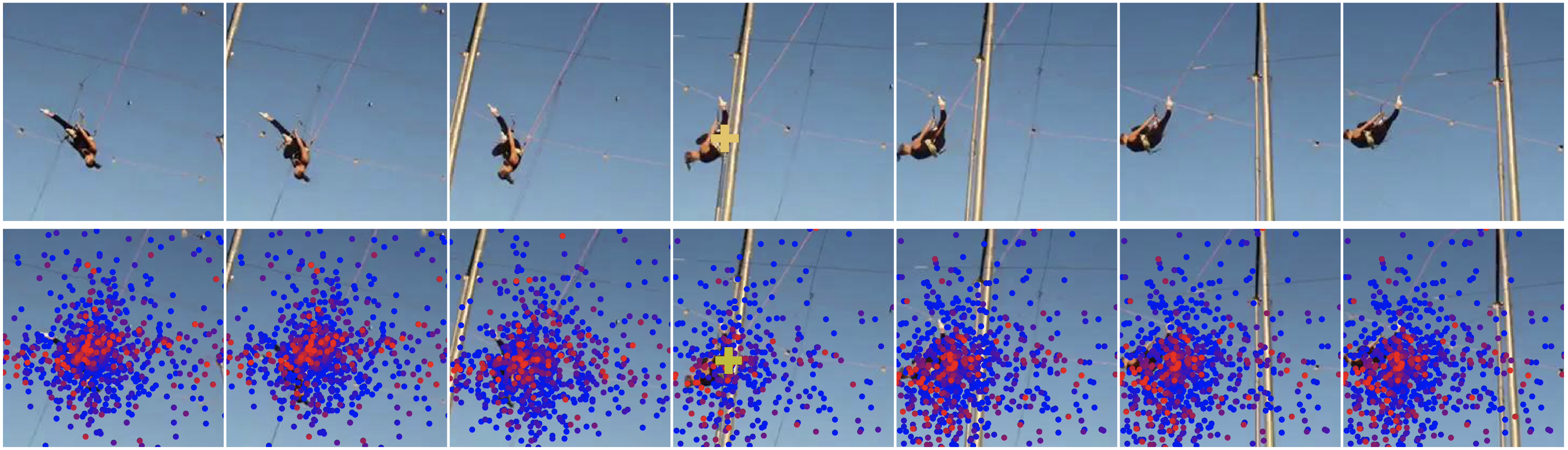}
\end{subfigure}
\caption{Visualization of patches attended by our deformable space-time attention. For each of the three examples, we show the original frames and, just below them, the frames with the sampled patches. The yellow cross in the middle frame shows the center of the query patch, while filled circles in each frame denote the patch centers sampled by our attention scheme for the given query. The color of the filled circles indicates the attention value assigned by our method (from blue to red). It can be seen that our attention strategy  harnesses the motion cues in the compressed video to successfully track the query object within the clip. The patches receiving highest weight values are those located on or near the human figure.}
\label{examples}
\end{figure*}

\section{Visualizations of Attended Patches}
In Figure~\ref{examples} we show the patches attended by our deformable space-time attention for three different sample sub-clips. The yellow cross in the middle frame denotes the query patch. Based on the appearance and the motion of the query patch, each attention layer of our model samples $N=8$ patches in each frame of the sub-clip. Filled circles denote sampled patch centers and their color (from blue to red) indicates the value
of the attention weights (from 0 to 1). We can see that, despite the presence of significant motion, our attention model can successfully track the query objects (the human subjects) and it concentrates the sampling in the most salient regions of the frames. The patches receiving highest weight values are those located on or near the query object.

\end{document}